\documentclass[conference]{IEEEtran}
\IEEEoverridecommandlockouts
\usepackage{amsmath,amssymb,amsfonts}
\usepackage{algorithmic}
\usepackage{algorithm}
\usepackage{graphicx}
\usepackage{textcomp}
\usepackage{booktabs}
\usepackage{multirow}
\usepackage{xcolor}
\usepackage{xurl}
\usepackage{hyperref}
\usepackage{tabularx}
\usepackage[utf8]{inputenc}
\usepackage{tikz}
\usetikzlibrary{arrows.meta,positioning,calc,fit,backgrounds}

\AddToHook{shipout/foreground}{%
  \ifnum\value{page}=1\relax
    \begin{tikzpicture}[remember picture,overlay]
      \node[anchor=south,align=center,text width=7.25in,font=\fontsize{5.1}{5.7}\selectfont]
        at ([yshift=0.08in]current page.south)
        {Accepted as a regular paper at IEEE ICMLA 2026.\\
        \textcopyright~2026 IEEE. Personal use of this material is permitted. Permission from IEEE must be obtained for all other uses, in any current or future media, including reprinting/republishing this material for advertising or promotional purposes, creating new collective works, for resale or redistribution to servers or lists, or reuse of any copyrighted component of this work in other works.};
    \end{tikzpicture}%
  \fi
}

\def\BibTeX{{\rm B\kern-.05em{\sc i\kern-.025em b}\kern-.08em
    T\kern-.1667em\lower.7ex\hbox{E}\kern-.125emX}}

\begin{document}

\title{Magnitude Profile Pruning: Calibration-Free Structured Attention Head Removal for Transformer Compression}

\author{
\IEEEauthorblockN{
Kasun Dewage\IEEEauthorrefmark{1},
Marianna Pensky\IEEEauthorrefmark{1},
Heranga K. Rathnasekara\IEEEauthorrefmark{2},
and Suranadi De Silva\IEEEauthorrefmark{1}
}

\IEEEauthorblockA{
\IEEEauthorrefmark{1}\textit{University of Central Florida}\\
Orlando, Florida, USA\\
\texttt{\{KasunTharuka.dewage, Marianna.Pensky, su966204\}@ucf.edu}
}

\IEEEauthorblockA{
\IEEEauthorrefmark{2}\textit{Old Dominion University}\\
Norfolk, Virginia, USA\\
\texttt{hrath001@odu.edu}
}
\thanks{Code is available at \url{https://github.com/Kasun-Dewage/Magnitude-Profile-Pruning-2026}.}
}

\maketitle

\begin{abstract}
Structured pruning of attention heads provides a hardware-friendly way to compress Transformer language models. However, existing methods for measuring head-level importance require calibration data, gradient computation, or Hessian estimation. These requirements add extra overhead and make the methods depend on the data. Our work presents \textit{Magnitude Profile (MP)} scoring, a training-free criterion for head importance that identifies dispensable heads through statistical outlier detection on weight row norms. Heads whose projection weights fall within the population bulk are pruned, while heads exhibiting outlier norms, which carry disproportionate representational capacity, are preserved. Our work further gives \textit{MP-G}, a variant that handles Grouped Query Attention (GQA) by distributing shared key-value group scores across associated query heads. Across five models  evaluated on WikiText-2 perplexity at 12.5\%--50\% head sparsity, MP-G achieves the best perplexity on OPT-6.7B at all sparsity levels (18.46 at 12.5\%, 27.87 at 25\%, 152.0 at 50\%). \textit{MP-G} also gives the best results on RoBERTa-large at 12.5\% and 25\%  sparsity, with perplexity values  of 7.27 and 10.28,  outperforming calibration-dependent baselines including Wanda-Head, SparseGPT-Head, and Gradient-Head. It requires \textit{zero} forward passes, calibration samples, or gradient computation. At 50\% sparsity, head pruning yields up to 16\% parameter reduction with 50\% attention FLOP savings. Our results show that weight-only statistical scoring can match or outperform data-dependent methods for structured head pruning, providing a practical, zero-cost criterion for Transformer compression.
\end{abstract}

\begin{IEEEkeywords}
Transformer pruning, attention heads, structured sparsity, model compression, language models
\end{IEEEkeywords}

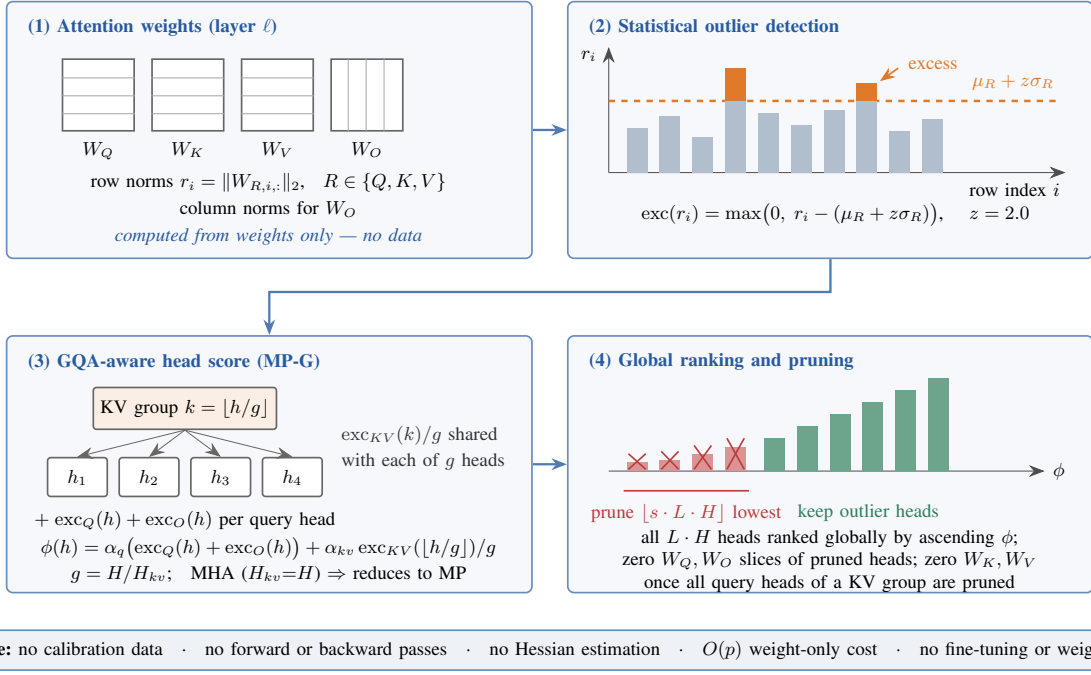
\begin{figure*}[!t]
\centering
\definecolor{mpblue}{RGB}{40,90,160}
\definecolor{mpbulk}{RGB}{176,190,205}
\definecolor{mpout}{RGB}{224,122,41}
\definecolor{mpkeep}{RGB}{46,125,90}
\definecolor{mpprune}{RGB}{190,60,60}
\resizebox{\textwidth}{!}{%
\begin{tikzpicture}[
  font=\small,
  >={Stealth[length=2.6mm]},
  stage/.style={draw=mpblue!70, thick, rounded corners=3pt, fill=mpblue!4, inner sep=0pt},
  stagetitle/.style={font=\small\bfseries, text=mpblue, anchor=west},
  wmat/.style={draw=black!60, thick, fill=white, minimum width=12mm, minimum height=12mm, inner sep=0pt},
  qhead/.style={draw=black!60, thick, fill=white, rounded corners=1.5pt, minimum width=10mm, minimum height=6.5mm, inner sep=1pt},
  flow/.style={->, line width=1.1pt, draw=mpblue!80}
]

%=========================================================
% STAGE 1: attention projection weights
%=========================================================
\begin{scope}[shift={(0,0)}]
  \node[stage, minimum width=8.8cm, minimum height=4.3cm, anchor=south west] (s1) at (0,0) {};
  \node[stagetitle] at (0.25,3.88) {(1) Attention weights (layer $\ell$)};

  \node[wmat] (wq) at (1.55,2.75) {};
  \foreach \y in {0.3,0.6,0.9} \draw[black!35] ($(wq.south west)+(0,\y)$) -- ($(wq.south east)+(0,\y)$);
  \node[below=0.5mm of wq] {$W_Q$};
  \node[wmat] (wk) at (3.05,2.75) {};
  \foreach \y in {0.3,0.6,0.9} \draw[black!35] ($(wk.south west)+(0,\y)$) -- ($(wk.south east)+(0,\y)$);
  \node[below=0.5mm of wk] {$W_K$};
  \node[wmat] (wv) at (4.55,2.75) {};
  \foreach \y in {0.3,0.6,0.9} \draw[black!35] ($(wv.south west)+(0,\y)$) -- ($(wv.south east)+(0,\y)$);
  \node[below=0.5mm of wv] {$W_V$};
  \node[wmat] (wo) at (6.05,2.75) {};
  \foreach \x in {0.3,0.6,0.9} \draw[black!35] ($(wo.south west)+(\x,0)$) -- ($(wo.north west)+(\x,0)$);
  \node[below=0.5mm of wo] {$W_O$};

  \node at (4.4,1.28) {row norms $r_i=\|W_{R,i,:}\|_2$, \; $R\in\{Q,K,V\}$};
  \node at (4.4,0.84) {column norms for $W_O$};
  \node[text=mpblue] at (4.4,0.38) {\emph{computed from weights only --- no data}};
\end{scope}

%=========================================================
% STAGE 2: outlier detection on the norm profile
%=========================================================
\begin{scope}[shift={(9.4,0)}]
  \node[stage, minimum width=8.8cm, minimum height=4.3cm, anchor=south west] (s2) at (0,0) {};
  \node[stagetitle] at (0.25,3.88) {(2) Statistical outlier detection};

  \draw[->, black!70] (0.7,1.45) -- (8.35,1.45);
  \node[black, anchor=north] at (7.5,1.40) {row index $i$};
  \draw[->, black!70] (0.7,1.45) -- (0.7,3.55);
  \node[black, anchor=east] at (0.62,3.40) {$r_i$};

  \draw[dashed, very thick, mpout] (0.7,2.65) -- (8.2,2.65);
  \node[text=mpout, anchor=south east] at (8.28,2.72) {$\mu_R+z\sigma_R$};

  \foreach \x/\h in {1.0/0.75, 1.55/0.95, 2.1/0.60, 3.2/1.00, 3.75/0.80, 4.3/1.05, 5.4/0.70, 5.95/0.90}
    \fill[mpbulk] (\x,1.45) rectangle ++(0.35,\h);
  \foreach \x/\h in {2.65/1.75, 4.85/1.50}{
    \fill[mpbulk] (\x,1.45) rectangle ++(0.35,1.20);
    \fill[mpout]  (\x,2.65) rectangle ++(0.35,\h-1.20);
  }
  \node[text=mpout, anchor=west] at (5.6,3.25) {excess};
  \draw[->, mpout, thick] (5.55,3.20) -- (5.24,2.98);

  \node at (4.5,0.75) {$\mathrm{exc}(r_i)=\max\bigl(0,\; r_i-(\mu_R+z\sigma_R)\bigr)$, \quad $z=2.0$};
\end{scope}

%=========================================================
% STAGE 3: GQA-aware head scores (MP-G)
%=========================================================
\begin{scope}[shift={(0,-5.6)}]
  \node[stage, minimum width=8.8cm, minimum height=4.3cm, anchor=south west] (s3) at (0,0) {};
  \node[stagetitle] at (0.25,3.88) {(3) GQA-aware head score (MP-G)};

  \node[draw=black!60, thick, fill=mpout!12, rounded corners=1.5pt,
        minimum width=27mm, minimum height=7mm] (kv) at (3.0,3.10)
        {KV group $k=\lfloor h/g\rfloor$};
  \foreach \i/\x in {1/1.2, 2/2.4, 3/3.6, 4/4.8}{
    \node[qhead] (q\i) at (\x,1.95) {$h_\i$};
    \draw[->, black!60] (kv.south) -- (q\i.north);
  }
  \node[anchor=west, text=black!80] at (5.5,2.65) {$\mathrm{exc}_{KV}(k)/g$ shared};
  \node[anchor=west, text=black!80] at (5.5,2.22) {with each of $g$ heads};
  \node at (3.0,1.22) {$+\;\mathrm{exc}_Q(h)+\mathrm{exc}_O(h)$ per query head};

  \node at (4.4,0.76) {$\phi(h)=\alpha_q\bigl(\mathrm{exc}_Q(h)+\mathrm{exc}_O(h)\bigr)+\alpha_{kv}\,\mathrm{exc}_{KV}(\lfloor h/g\rfloor)/g$};
  \node at (4.4,0.32) {$g=H/H_{kv}$;\quad MHA ($H_{kv}{=}H$) $\Rightarrow$ reduces to MP};
\end{scope}

%=========================================================
% STAGE 4: global ranking and pruning
%=========================================================
\begin{scope}[shift={(9.4,-5.6)}]
  \node[stage, minimum width=8.8cm, minimum height=4.3cm, anchor=south west] (s4) at (0,0) {};
  \node[stagetitle] at (0.25,3.88) {(4) Global ranking and pruning};

  \draw[->, black!70] (0.7,2.05) -- (8.0,2.05);
  \node[black, anchor=west] at (8.02,2.05) {$\phi$};
  \foreach \x/\h in {1.0/0.06, 1.55/0.10, 2.1/0.20, 2.65/0.32}{
    \fill[mpprune!55] (\x,2.05) rectangle ++(0.35,\h+0.08);
    \draw[mpprune, thick] (\x+0.03,2.08) -- ++(0.29,\h+0.20);
    \draw[mpprune, thick] (\x+0.32,2.08) -- ++(-0.29,\h+0.20);
  }
  \foreach \x/\h in {3.3/0.55, 3.85/0.75, 4.4/0.95, 4.95/1.15, 5.5/1.35, 6.05/1.55}
    \fill[mpkeep!70] (\x,2.05) rectangle ++(0.35,\h);
  \draw[mpprune, thick] (0.95,1.72) -- (3.05,1.72);
  \node[text=mpprune, anchor=north] at (2.0,1.62) {prune $\lfloor s\cdot L\cdot H\rfloor$ lowest};
  \node[text=mpkeep, anchor=north] at (5.05,1.62) {keep outlier heads};

  \node at (4.4,0.92) {all $L\cdot H$ heads ranked globally by ascending $\phi$;};
  \node at (4.4,0.54) {zero $W_Q,W_O$ slices of pruned heads; zero $W_K,W_V$};
  \node at (4.4,0.16) {once all query heads of a KV group are pruned};
\end{scope}

% arrows between stages
\draw[flow] (s1.east) -- (s2.west);
\draw[flow] (s2.south) -- ++(0,-0.55) -| (s3.north);
\draw[flow] (s3.east) -- (s4.west);

% bottom banner
\node[draw=mpblue!70, thick, rounded corners=3pt, fill=mpblue!8, inner sep=5pt,
      minimum width=18.2cm, align=center, font=\small]
  at (9.1,-6.55)
  {\textbf{Calibration-free:} no calibration data \; $\cdot$ \; no forward or backward passes \; $\cdot$ \;
   no Hessian estimation \; $\cdot$ \; $O(p)$ weight-only cost \; $\cdot$ \; no fine-tuning or weight reconstruction};

\end{tikzpicture}%
}
\caption{Overview of Magnitude Profile Pruning (MP-G). (1)~For each layer, L2 row norms are computed for the $W_Q, W_K, W_V$ projections and column norms for $W_O$, using model weights only. (2)~Per projection, the excess of each norm beyond the statistical threshold $\mu_R + z\sigma_R$ (Eq.~\ref{eq:excess}, $z{=}2.0$ by default) identifies outlier rows; norms within the population bulk receive zero excess. (3)~Head scores aggregate query-side excesses ($\mathrm{exc}_Q + \mathrm{exc}_O$) and, for GQA models, distribute the shared KV-group excess equally across the $g = H/H_{kv}$ query heads in the group (Eq.~\ref{eq:mpg}); for MHA models MP-G reduces to standard MP. (4)~Heads are ranked globally by ascending score and the $\lfloor s \cdot L \cdot H \rfloor$ lowest-scoring heads are pruned by zeroing their $W_Q$ and $W_O$ slices, with $W_K, W_V$ zeroed once all query heads in a KV group are pruned (Algorithm~\ref{alg:mpg}). The entire procedure requires no calibration samples, forward passes, gradients, or Hessian estimation.}
\label{fig:overview}
\end{figure*}

\section{Introduction}

Large Transformer-based language models~\cite{vaswani2017attention} deliver state-of-the-art performance across natural language tasks, but their computational and memory footprints challenge deployment in resource-constrained settings. Pruning is a well-established strategy for reducing model size~\cite{lecun1990optimal}, yet most work has targeted \textit{unstructured} weight pruning~\cite{frantar2023sparsegpt,sun2023wanda}. This approach achieves fine-grained sparsity but requires specialized hardware or software to convert sparsity into wall-clock speedup.

\textit{Structured} pruning at the attention head level is more hardware-friendly: entire head weight slices can be zeroed, directly reducing the projection operations and attention computations. However, existing head-level importance criteria carry non-trivial overhead. Learned-gate approaches~\cite{voita2019analyzing} require training with $L_0$ regularization. Gradient-based criteria~\cite{michel2019sixteen} and Taylor-based methods~\cite{molchanov2017pruning} require backward passes over calibration data. Even our adaptations of Wanda~\cite{sun2023wanda} and SparseGPT~\cite{frantar2023sparsegpt} to head granularity require forward passes and activation/Hessian collection over calibration samples.

We argue that the weight matrices themselves contain sufficient signal to identify dispensable heads, without any data-dependent computation. Our key observation is that, in trained Transformers, a small subset of attention heads develops projection weights with significantly larger norms than the population average. These \textit{outlier heads} carry more information, while heads with typical, near-average norms are mostly redundant and can be safely removed.

Building on this insight, we propose \textbf{Magnitude Profile (MP)} scoring, which quantifies head importance via the excess of weight row norms beyond a statistical threshold $(\mu + z\sigma)$. Heads with no excess, i.e.\ those whose weights are statistically typical, receive a score of zero and are pruned first. We further introduce \textbf{MP-G}, a variant designed for modern Grouped Query Attention (GQA) architectures by computing shared key-value group scores and distributing them proportionally across associated query heads.

Our contributions are:\\
% \begin{itemize}
% \item
\hspace*{2mm} 1.\ We introduce \textbf{Magnitude Profile (MP)} scoring, a calibration-free, gradient-free head importance criterion based on statistical outlier detection in weight norms. The method requires only model weight access, with $O(p)$ computation where $p$ is the number of attention parameters.\\
%
% \item 
\hspace*{2mm} 2.\ We propose \textbf{MP-G}, a GQA-aware extension that accounts for key-value head sharing in modern architectures such as LLaMA-3 and Mistral.\\
%
%\item 
\hspace*{2mm} 3.\  We demonstrate through experiments on \textbf{five models} across encoder and decoder architectures that MP-G achieves the best perplexity in several settings and top-two performance in most primary model-sparsity settings at 12.5\%--50\% head sparsity, while requiring zero calibration overhead.\\
%
% \item 
\hspace*{2mm} 4.\ We provide efficiency analysis showing 50\% head pruning yields approximately 7--16\% total parameter reduction with 50\% attention FLOP savings.
% \end{itemize}

\section{Related Work}

\textbf{Attention Head Pruning.}
Michel et al.~\cite{michel2019sixteen} showed that many attention heads can be removed with minimal accuracy loss and proposed gradient-based importance scoring via Taylor expansion. Voita et al.~\cite{voita2019analyzing} employed differentiable gates with $L_0$ regularization to identify and prune redundant heads during training. Both methods require backward passes over task-specific data, limiting their applicability to settings where calibration data and training infrastructure are available.

\textbf{Unstructured Pruning Criteria.}
Magnitude pruning, which removes weights with smallest absolute values, remains a strong baseline~\cite{han2015learning}. SparseGPT~\cite{frantar2023sparsegpt} uses approximate Hessian information to prune and reconstruct weights row-by-row with a single forward pass. Wanda~\cite{sun2023wanda} combines weight magnitude with input activation norms, achieving competitive unstructured sparsity without weight updates. Both were designed for element-level sparsity, and their adaptation to structured head-level pruning has not been systematically studied.

\textbf{Structured Pruning of LLMs.}
LLM-Pruner~\cite{ma2023llmpruner} uses gradient information to identify coupled structures for removal. Sheared LLaMA~\cite{xia2023sheared} learns pruning masks jointly with continued pre-training. SliceGPT~\cite{ashkboos2024slicegpt} removes entire rows and columns via PCA-based projections. CoFi~\cite{xia2022cofi} performs coarse-grained and fine-grained pruning jointly, targeting both attention heads and feed-forward layers with learned masks. These approaches target broader structural units (layers, hidden dimensions) and typically require substantial computation. Our work focuses on head-level pruning using a lightweight, training-free criterion.

\section{Method}

\subsection{Problem Formulation}

Consider a Transformer~\cite{vaswani2017attention} with $L$ layers, each containing $H$ attention heads of dimension $d_h = d_{\text{model}} / H$. Each head $h$ in layer $\ell$ is parameterized by projection matrices $W_Q^{(\ell,h)}, W_K^{(\ell,h)}, W_V^{(\ell,h)} \in \mathbb{R}^{d_h \times d_{\text{model}}}$ and output projection columns $W_{O,\, h \cdot d_h : (h+1) \cdot d_h}^{(\ell)} \in \mathbb{R}^{d_{\text{model}} \times d_h}$.

Given a target head sparsity ratio $s \in (0,1)$, the goal is to identify and zero the Q, K, V, and O weight slices of $\lfloor s \cdot L \cdot H \rfloor$ heads globally. Physical removal of pruned head dimensions (model surgery) is left for deployment; our experiments zero head slices in place. This process requires a scoring function $\phi(\ell, h) \in \mathbb{R}_{\geq 0}$ that assigns importance to each head; heads with the lowest scores are pruned.

\subsection{Magnitude Profile Scoring}

The central idea behind MP scoring is that trained Transformers exhibit heterogeneous weight norm distributions across heads. Rather than treating raw weight magnitude as the importance signal, which conflates overall weight scale with head-specific significance, MP identifies heads that are \textit{statistically indistinguishable} from the population as redundant, and preserves only those with \textit{outlier} weight patterns.

For each projection matrix $R \in \{Q, K, V\}$ of a given layer, we compute the L2 row norms $r_i = \|W_{R,i,:}\|_2$ for all rows $i$. We then define the excess beyond a threshold controlled by a parameter $z$:
\begin{equation}
\text{exc}(r_i) = \max\!\big(0,\; r_i - (\mu_r + z \cdot \sigma_r)\big)
\label{eq:excess}
\end{equation}
where $\mu_r$ and $\sigma_r$ are the mean and standard deviation of all row norms in that projection matrix. For the output projection $W_O$, column norms are used analogously, since $W_O$ is transposed relative to Q/K/V in terms of head indexing.

The head-level importance score aggregates the excesses across all four projection roles:
\begin{equation}
\phi_{\text{MP}}(\ell, h) = \sum_{R \in \{Q,K,V,O\}} \sum_{i \in \text{rows}(h,R)} \text{exc}(r_i^R)
\label{eq:mp}
\end{equation}

The threshold $z$ controls the sensitivity of outlier detection. A high $z$ (e.g., 2.5) preserves only extreme outliers and aggressively prunes ``normal'' heads. A lower $z$ (e.g., 1.5) is more conservative and retains heads with moderately elevated norms. We use $z=2.0$ as the default, following the common two-standard-deviation rule.

\textbf{Intuition.} If all heads in a layer have similar weight norms, they contribute roughly equally and some can be removed with limited damage. But if a few heads have much larger norms, those heads likely carry specialized functions (e.g., positional attention, syntactic parsing) that would be destructive to remove~\cite{voita2019analyzing,clark2019does}. MP formalizes this idea by assigning zero importance to heads within the bulk distribution and positive importance only to outliers.

\textbf{Computational cost.} MP requires a single pass over all attention weight matrices to compute norms and statistics. For a model with $p$ attention parameters, the computational cost is $O(p)$ with no forward passes, no backward passes, no calibration data, and no Hessian estimation.

\subsection{MP-G: GQA-Aware Extension}

Modern decoder models such as LLaMA-3~\cite{dubey2024llama3} and Mistral~\cite{jiang2023mistral} use Grouped Query Attention (GQA)~\cite{ainslie2023gqa}, where $H_{kv} < H$ key-value heads are shared across groups of $g = H / H_{kv}$ query heads. Standard MP applies the same K/V excess to all query heads in a group, failing to account for the structural coupling.

MP-G addresses this by computing a per-KV-group score and distributing it equally across associated query heads:
\begin{equation}
\phi_{\text{MP-G}}(h) = \alpha_q \cdot \big(\text{exc}_Q(h) + \text{exc}_O(h)\big) + \alpha_{kv} \cdot \frac{\text{exc}_{KV}(\lfloor h/g \rfloor)}{g}
\label{eq:mpg}
\end{equation}
where $\text{exc}_{KV}(k) = \sum_{i \in \text{rows}(k)} \big(\text{exc}(r_i^K) + \text{exc}(r_i^V)\big)$ is the total excess for KV group $k$, and $\alpha_q, \alpha_{kv}$ control the relative weighting of query-side vs.\ KV-side contributions. We use $\alpha_q = \alpha_{kv} = 1.0$ by default.

The division by $g$ ensures that the shared KV signal is not over-counted across the group. When all query heads in a KV group are pruned, the corresponding K/V weights are also zeroed, since no query head references them.

For Multi-Head Attention (MHA) models where $H_{kv} = H$, MP-G reduces to standard MP.

\subsection{Pruning Procedure}

Algorithm~\ref{alg:mpg} summarizes the complete pruning procedure. Given a model and target sparsity $s$, we (1) compute MP-G scores for all heads across all layers, (2) globally rank heads by ascending score, and (3) zero the weight slices of the lowest-scoring heads until the target sparsity is reached. No fine-tuning or weight reconstruction is performed.

\begin{algorithm}[t]
\caption{MP-G Head Pruning}
\label{alg:mpg}
\begin{algorithmic}[1]
\REQUIRE Model $M$ with $L$ layers, $H$ heads/layer, $H_{kv}$ KV heads/layer, target sparsity $s$, threshold $z$
\STATE $g \leftarrow H / H_{kv}$ \COMMENT{GQA group size}
\STATE $\mathit{scores} \leftarrow []$
\FOR{each layer $\ell = 1, \ldots, L$}
  \FOR{each projection $R \in \{Q, K, V, O\}$}
    \STATE Compute row norms for $Q,K,V$ and column norms for $O$; compute $\mu^R, \sigma^R$
    \STATE $\text{exc}^R_i \leftarrow \max(0, r^R_i - \mu^R - z \cdot \sigma^R) \;\; \forall i$
  \ENDFOR
  \FOR{each KV group $k = 1, \ldots, H_{kv}$}
    \STATE $e_{kv}(k) \leftarrow \sum_{i \in \text{group}(k)} (\text{exc}^K_i + \text{exc}^V_i)$
  \ENDFOR
  \FOR{each query head $h = 1, \ldots, H$}
    \STATE $e_q(h) \leftarrow \sum_{i \in \text{head}(h)} (\text{exc}^Q_i + \text{exc}^O_i)$
    \STATE $\phi(\ell, h) \leftarrow \alpha_q \cdot e_q(h) + \alpha_{kv} \cdot e_{kv}(\lfloor h/g \rfloor) / g$
    \STATE Append $(\ell, h, \phi)$ to $\mathit{scores}$
  \ENDFOR
\ENDFOR
\STATE Sort $\mathit{scores}$ by $\phi$ ascending
\STATE $n_{\text{prune}} \leftarrow \lfloor s \cdot L \cdot H \rfloor$
\FOR{$i = 1, \ldots, n_{\text{prune}}$}
  \STATE Zero $W_Q^{(\ell,h)}, W_O^{(\ell,h)}$ slices for head $(\ell, h)$
  \IF{all query heads in KV group $\lfloor h/g \rfloor$ are pruned}
    \STATE Zero $W_K^{(\ell,k)}, W_V^{(\ell,k)}$ slices
  \ENDIF
\ENDFOR
\RETURN Pruned model $M'$
\end{algorithmic}
\end{algorithm}

\subsection{Baseline Scoring Methods}

We compare MP-G against four head-level scoring criteria adapted from the pruning literature.

\textbf{Wanda-Head.} Adapts the Wanda criterion~\cite{sun2023wanda} to head granularity by aggregating the product of weight magnitudes and input activation norms across Q/K/V rows and O columns of each head:
\begin{equation}
\begin{aligned}
\phi_{\text{Wanda}}(h) ={}& \sum_{R \in \{Q,K,V\}} \sum_{i \in \text{rows}_R(h)} \sum_j |W_{R,ij}| \cdot \|X^{\mathrm{in}}_j\|_2 \\
&+ \sum_{j \in \text{cols}_O(h)} \sum_i |W_{O,ij}| \cdot \|X^{O}_j\|_2.
\end{aligned}
\end{equation}
This requires forward passes over calibration data to collect activation statistics.

\textbf{SparseGPT-Head.} Aggregates a SparseGPT-inspired Hessian-diagonal importance score $w_{ij}^2 \cdot H_{jj}$~\cite{frantar2023sparsegpt} \underline{at} the head level, where $H_{jj}$ is the diagonal of the empirical Hessian of the layer input. This requires calibration data and Hessian estimation.

\textbf{Gradient-Head.} Uses the first-order criterion $|w_{ij} \cdot g_{ij}|$~\cite{michel2019sixteen,molchanov2017pruning}, where $g_{ij}$ is the gradient of the language modeling loss with respect to each weight, aggregated per head. This requires a backward pass over calibration data.

\textbf{Hybrid.} Normalizes Magnitude, MP, and Wanda scores to $[0,1]$ via min-max scaling, then combines: $\phi_{\text{Hyb}} = \lambda_{\text{mag}} \hat{s}_{\text{mag}} + \lambda_{\text{mp}} \hat{s}_{\text{mp}} + \lambda_{\text{wanda}} \hat{s}_{\text{wanda}}$ with $\lambda_{\text{mag}} = \lambda_{\text{mp}} = \lambda_{\text{wanda}} = 1.0$. This tests whether fusing weight-only and calibration signals improves over either alone.

\section{Experimental Setup}

\textbf{Models.} We evaluate five pre-trained models spanning encoder and decoder architectures: RoBERTa-base and RoBERTa-large~\cite{liu2019roberta} (125M, 12 heads $\times$ 12 layers; 355M, 16 $\times$ 24), OPT-6.7B~\cite{zhang2022opt} (32 $\times$ 32, MHA), LLaMA-3-8B~\cite{dubey2024llama3} (32 $\times$ 32, GQA with 8 KV heads), and Mistral-7B-v0.1~\cite{jiang2023mistral} (32 $\times$ 32, GQA with 8 KV heads).

\textbf{Sparsity Levels.} We evaluate at 12.5\%, 25\%, and 50\% attention head sparsity, applied uniformly via global ranking across layers.

\textbf{Perplexity Evaluation.} For decoder models, we report causal language modeling perplexity on the WikiText-2~\cite{merity2016pointer} validation split (FP16). For encoder models (RoBERTa), we report masked language modeling pseudo-perplexity (FP32).

\textbf{GLUE Evaluation.} For all reported GLUE runs, we fine-tune RoBERTa-large on SST-2 (accuracy), MRPC (F1), and RTE (accuracy)~\cite{wang2019glue} and then apply head pruning before evaluation (fine-tune-then-prune protocol), with learning rate $10^{-5}$ and batch size 16 for 3 epochs (SST-2), 5 epochs (MRPC), and 10 epochs (RTE).

\textbf{Calibration.} Methods requiring calibration data use WikiText-2 training samples with a sequence length of 512. Wanda-Head, SparseGPT-Head, and Hybrid use 64 samples, while Gradient-Head uses 32 samples for gradient collection. \textbf{MP and MP-G require no calibration data.}

\textbf{Hyperparameters.} Default threshold $z{=}2.0$ for MP/MP-G. Hybrid uses equal combination weights $\lambda{=}1.0$. All experiments use seed 0.

\section{Results}

\subsection{Perplexity Across Models}

Table~\ref{tab:ppl_all} presents perplexity results across all five models at three sparsity levels. For encoder models (RoBERTa), we report pseudo-perplexity and for decoder models, we report causal language modeling perplexity on WikiText-2.

\begin{table*}[t]
\centering
\caption{Perplexity ($\downarrow$) on WikiText-2 at head sparsity $s$ with $z{=}2.0$. RoBERTa models report pseudo-perplexity (FP32); decoder models report causal LM perplexity (FP16). ``--'' indicates unavailable result. Best in \textbf{bold}, second \underline{underlined}.}
\label{tab:ppl_all}
\setlength{\tabcolsep}{3pt}
\small
\begin{tabular}{l|ccc|ccc|ccc|ccc|ccc}
\toprule
 & \multicolumn{3}{c|}{RoBERTa-base} & \multicolumn{3}{c|}{RoBERTa-large} & \multicolumn{3}{c|}{OPT-6.7B} & \multicolumn{3}{c|}{LLaMA-3-8B} & \multicolumn{3}{c}{Mistral-7B} \\
Method & 12.5 & 25 & 50 & 12.5 & 25 & 50 & 12.5 & 25 & 50 & 12.5 & 25 & 50 & 12.5 & 25 & 50 \\
\midrule
Grad   & \textbf{8.22} & \textbf{11.8} & \textbf{64.7} & \underline{7.79} & 81.8 & 13485 & 28.5 & -- & -- & \textbf{11.1} & \textbf{17.7} & \textbf{47.9} & 93.6 & \textbf{13.3} & \textbf{54.1} \\
Wanda  & \underline{8.89} & 16.5 & 103.8 & 8.01 & 13.7 & 86.0 & 42.5 & 152.3 & \underline{406.3} & 1608 & -- & 20009 & 107.0 & 472.6 & \underline{552.8} \\
SpGPT  & 11.2 & 113.4 & 4730 & 7.90 & \underline{10.3} & \textbf{42.2} & 257.9 & 479.0 & 2034 & 142.3 & 344.6 & \underline{1870} & 75.3 & 430.8 & 871.8 \\
MP-G   & 10.0 & 17.8 & \underline{71.3} & \textbf{7.27} & \textbf{10.3} & \underline{45.8} & \textbf{18.5} & \textbf{27.9} & \textbf{152.0} & \underline{20.4} & \underline{41.4} & 7227 & \underline{22.5} & \underline{121.0} & 1072 \\
Hybrid & 9.09 & \underline{14.0} & 96.1 & 7.82 & 11.1 & 85.6 & \underline{25.6} & \underline{136.3} & 499.6 & 40.3 & 467.9 & 13428 & \textbf{14.3} & 1290 & 4339 \\
\bottomrule
\end{tabular}
\end{table*}

On RoBERTa-large, MP-G achieves the lowest perplexity at 12.5\% sparsity (7.27) and 25\% sparsity (10.28), outperforming all calibration-dependent methods. At 50\%, MP-G (45.76) is close to the best method, SparseGPT-Head (42.23), while requiring no Hessian computation. The gap of only 3.5 perplexity points at this aggressive sparsity level is notable given MP-G's zero calibration cost.

On RoBERTa-base, Gradient-Head performs best at all sparsity levels. This suggests that gradient information is particularly valuable for smaller models where weight-norm heterogeneity may be less pronounced. MP-G ranks second at 50\% sparsity (71.25 vs.\ 64.74), and outperforms Wanda-Head (103.8) and SparseGPT-Head (4730) by a wide margin.

MP-G performs best on OPT-6.7B, obtaining the lowest perplexity at all three sparsity levels: 18.46 (12.5\%), 27.87 (25\%), and 152.0 (50\%). At 12.5\%, MP-G improves upon the second-best method (Hybrid, 25.64) by 28\%. At 50\%, MP-G reduces perplexity by 63\% relative to Wanda-Head (406.3) and by 93\% relative to SparseGPT-Head (2034). These results indicate that the weight-norm heterogeneity in OPT-6.7B is highly informative for head selection. On OPT-6.7B, Gradient-Head results at 25\% and 50\% sparsity are unavailable in our aggregate results, preventing direct comparison at higher pruning ratios.

On LLaMA-3-8B, Gradient-Head achieves the best results across all sparsity levels (11.12 at 12.5\%, 17.69 at 25\%, 47.94 at 50\%), while MP-G ranks second at 12.5\% (20.39) and 25\% (41.41). On Mistral-7B, MP-G performs competitively at low sparsity (22.48 at 12.5\%), while Gradient-Head leads at 25\% (13.30) and 50\% (54.11). The Hybrid method achieves 14.27 at 12.5\% on Mistral, its best result across all models. At 50\% sparsity on Mistral, among the non-gradient baselines, Wanda-Head (552.8) outperforms both MP-G (1072) and SparseGPT-Head (871.8), suggesting that activation-based signals can be important in certain GQA architectures at high sparsity. Gradient-Head remains best in this setting.

\subsection{Effect of the Threshold $z$}

The MP threshold $z$ controls outlier sensitivity: higher $z$ preserves fewer, more extreme outlier heads. Table~\ref{tab:z_ablation} shows its effect on selected model-sparsity pairs.

\begin{table}[t]
\centering
\caption{Effect of threshold $z$ on MP-G perplexity at 25\% and 50\% sparsity.}
\label{tab:z_ablation}
\setlength{\tabcolsep}{4pt}
\begin{tabular}{l|cc|cc}
\toprule
 & \multicolumn{2}{c|}{$s{=}25\%$} & \multicolumn{2}{c}{$s{=}50\%$} \\
Model & $z{=}1.5$ & $z{=}2.0$ & $z{=}2.0$ & $z{=}2.5$ \\
\midrule
RoBERTa-base & \textbf{10.67} & 17.79 & \textbf{71.25} & 73.67 \\
RoBERTa-large & 10.45 & \textbf{10.28} & 45.76 & \textbf{41.19} \\
LLaMA-3-8B & 54.98 & \textbf{41.41} & 7227 & \textbf{282.8} \\
Mistral-7B & \textbf{40.09} & 121.0 & \textbf{1072} & 2479 \\
\bottomrule
\end{tabular}
\end{table}

Two patterns are observed. First, the optimal $z$ is model-dependent: MHA models (RoBERTa-base) often prefer lower $z$ (1.5), while GQA models (LLaMA-3) and larger encoders (RoBERTa-large) benefit from higher $z$ (2.0--2.5). This suggests that GQA architectures, where fewer KV heads create stronger outlier structure, benefit from more aggressive outlier-based filtering. Second, $z{=}2.5$ produces a large improvement on LLaMA-3-8B at 50\% sparsity (from 7227 to 282.8), indicating that at extreme pruning ratios, preserving only the most extreme outliers is critical. These results suggest that a model-adaptive $z$ selection strategy could further improve MP-G, which we leave to future work.

\subsection{Downstream Task Performance}

Table~\ref{tab:glue} reports GLUE~\cite{wang2019glue} results for RoBERTa-large with the fine-tune-then-prune protocol. At 25\% sparsity, all reported methods retain substantial task performance on SST-2 (92.8--95.2\% accuracy). SparseGPT-Head shows the strongest downstream resilience overall, maintaining 89.6\% SST-2 accuracy and 81.4 MRPC F1 even at 50\% sparsity, where MP-G, Wanda-Head, and Hybrid collapse on MRPC. The Hybrid method achieves the highest SST-2 accuracy at 25\% sparsity (95.2\%).

The gap between perplexity and downstream performance at 50\% sparsity is worth examining: MP-G achieves competitive perplexity (45.76 vs.\ 42.23 for SparseGPT-Head), yet SparseGPT-Head shows better downstream robustness after post-fine-tuning pruning. This suggests that Hessian-guided scoring better preserves the task-specific representations learned during fine-tuning, whereas MP-G preserves heads important for general language modeling. At 50\% sparsity on RTE, all reported methods converge to near-chance accuracy (47.3\%), indicating this challenging task cannot tolerate removal of half its attention capacity.

\begin{table}[t]
\centering
\caption{GLUE results for RoBERTa-large (fine-tune-then-prune). SST-2 and RTE: accuracy; MRPC: F1.}
\label{tab:glue}
\setlength{\tabcolsep}{3pt}
\small
\begin{tabular}{l|cc|cc|cc}
\toprule
 & \multicolumn{2}{c|}{SST-2 (Acc)} & \multicolumn{2}{c|}{MRPC (F1)} & \multicolumn{2}{c}{RTE (Acc)} \\
Method & 25\% & 50\% & 25\% & 50\% & 25\% & 50\% \\
\midrule
Wanda  & 93.5 & 81.4 & 28.8 & 0.0 & \underline{51.6} & 47.3 \\
SpGPT  & \underline{94.6} & \textbf{89.6} & \textbf{88.6} & \textbf{81.4} & \textbf{75.1} & 47.3 \\
MP-G   & 92.8 & 82.9 & \underline{79.7} & 0.0 & 48.4 & 47.3 \\
Hybrid & \textbf{95.2} & \underline{86.0} & 60.4 & 0.0 & 50.9 & 47.3 \\
\bottomrule
\end{tabular}
\end{table}

\subsection{Efficiency Analysis}

At 50\% head sparsity, all scoring methods remove the same number of attention heads, so the theoretical sparsity-dependent efficiency effects are identical across methods.

Parameter reduction ranges from 8\% for GQA models (where KV parameters are shared and MLP parameters dominate) to 16\% for MHA models. All configurations achieve the theoretical 50\% reduction in attention FLOPs. However, measured wall-clock change is modest, ranging from small slowdowns to about 3.5\% speedup, because standard PyTorch attention kernels do not dynamically skip zeroed heads. Realizing the theoretical FLOP savings requires either model surgery to physically remove pruned head dimensions or integration with sparse inference engines, an important engineering step for deployment.

\section{Analysis}

\subsection{Why Does MP-G Work?}

MP-G's effectiveness rests on the empirical observation that trained Transformer attention weights are not uniformly distributed across heads. Instead, a minority of heads develop disproportionately large weight norms, an outlier structure that MP exploits. This observation is consistent with prior findings that large-magnitude features emerge in trained Transformers~\cite{dettmers2022llmint8,kovaleva2019revealing}. On OPT-6.7B and RoBERTa-large, where this heterogeneity is most pronounced, MP-G outperforms all baselines. On LLaMA-3-8B, where Gradient-Head performs best, the weight norm distribution may be more uniform, reducing MP's discriminative power.

This heterogeneity has been independently observed in other contexts: Voita et al.~\cite{voita2019analyzing} found that only a few ``specialized'' heads perform critical functions (positional, syntactic, rare-token attention), while the majority are redundant. Clark et al.~\cite{clark2019does} showed that BERT attention heads exhibit distinct patterns, with some heads attending broadly while others focus on specific linguistic relations. MP-G can be understood as detecting these specialized heads via their weight-norm signatures, without needing to observe their attention patterns on data.

\subsection{Calibration-Free Advantage}

A practical strength of MP-G is its complete independence from data. The method:
\begin{itemize}
\item Requires \textbf{no calibration dataset}, eliminating distributional mismatch concerns, which is particularly relevant when the pruning-time domain differs from the deployment domain.
\item Requires \textbf{no forward or backward passes}, making it applicable even when GPU memory is insufficient to run inference on the full model.
\item Executes in \textbf{seconds} (weight norm computation) rather than minutes (forward passes over 64 calibration sequences of length 512), making it suitable for rapid prototyping and hyperparameter sweeps.
\end{itemize}

In contrast, Wanda-Head, SparseGPT-Head, and Gradient-Head each require 32--64 calibration sequences, forward passes to collect activations or Hessians, and (for Gradient-Head) backward passes. For a 7B-parameter model, this overhead is non-negligible and scales with calibration set size.

\subsection{Method Robustness Across Architectures}

Table~\ref{tab:wins} summarizes competitive performance by counting the number of model-sparsity combinations where each method achieves the best or second-best perplexity. We only consider results at $z{=}2.0$ where at least two methods produce finite perplexity. All 15 model-sparsity settings satisfy this condition.

\begin{table}[t]
\centering
\caption{Number of model-sparsity pairs where each method achieves best or second-best perplexity (out of 15 comparable pairs at $z{=}2.0$ with ${\geq}2$ finite results).}
\label{tab:wins}
\setlength{\tabcolsep}{4pt}
\begin{tabular}{l|cc|c}
\toprule
Method & Best & 2nd & Calib.\ Required \\
\midrule
Grad       & 8  & 1 & Yes (backward) \\
MP-G       & 5  & 6 & No \\
SpGPT      & 1  & 2 & Yes (Hessian) \\
Hybrid     & 1  & 3 & Yes (forward) \\
Wanda      & 0  & 3 & Yes (forward) \\
\bottomrule
\end{tabular}
\end{table}

Gradient-Head achieves the most first-place finishes, with eight. However, its OPT-6.7B results at 25\% and 50\% sparsity are unavailable in our aggregate results. MP-G achieves top-two performance in most primary model-sparsity settings and produces finite perplexity on every tested configuration. This combination of competitive performance and broad robustness across architectures is a useful property when calibration data is unavailable.

\section{Conclusion}

We have introduced Magnitude Profile (MP-G) scoring, a calibration-free structured attention head pruning criterion based on weight-norm outlier detection. MP-G requires access only to model weights. It does not require calibration data, forward passes, gradients, or Hessian estimation, while still achieving competitive or superior perplexity in several settings and top-two performance in most primary model-sparsity settings. On OPT-6.7B, MP-G leads at all sparsity levels; on RoBERTa-large, it leads at 12.5\% and 25\% and closely matches the best Hessian-based method at 50\%.

For practitioners, MP-G offers a fast, zero-cost initial pruning strategy: compute head scores in seconds, prune, and optionally fine-tune if downstream task performance is critical. The method is particularly suited to scenarios where calibration data is unavailable, domain-mismatched, or where rapid pruning iteration is needed. Future work should explore automatic selection of $z$, integration with MLP pruning for deeper compression, and model surgery or sparse kernel support to translate the theoretical FLOP savings into measured speedup.

\bibliographystyle{IEEEtran}

\end{document}